\documentclass[journal=jacsat,manuscript=article]{achemso}

\usepackage[version=3]{mhchem} 
\usepackage{amssymb} 

\author{Ziyue Wang}
\affiliation[CMU]
{Department of Materials Science and Engineering, Carnegie Mellon University, Pittsburgh, PA, USA}

\author{Yayati Jadhav}
\affiliation[CMU]
{Department of Mechanical Engineering, Carnegie Mellon University, Pittsburgh, PA, USA}

\author{Peter Pak}
\affiliation[CMU]
{Department of Mechanical Engineering, Carnegie Mellon University, Pittsburgh, PA, USA}

\author{Amir Barati Farimani}
\email{barati@cmu.edu}
\affiliation[CMU]
{Department of Mechanical Engineering, Carnegie Mellon University, Pittsburgh, PA, USA}

\title[An \textsf{achemso} demo]
  {Image2Gcode: Image-to-G-code Generation for Additive Manufacturing Using Diffusion-Transformer Model}

\abbreviations{IR,NMR,UV}

\begin{document}



\begin{abstract}

Mechanical design and manufacturing workflows conventionally begin with conceptual design, followed by the creation of a detailed computer-aided design (CAD) model and fabrication through material-extrusion (MEX) printing. This process requires converting CAD geometry into machine-readable G-code through slicing and path planning. While each step is well established, the dependence on CAD modeling remains a major bottleneck: constructing object-specific 3D geometry is slow, expertise-intensive, and poorly suited to rapid or ad hoc prototyping scenarios. Even minor design variations typically necessitate manual updates in CAD software, making iteration time-consuming, designer-dependent, and difficult to scale.
To address this limitation, we introduce Image2Gcode, an end-to-end data-driven framework that bypasses the CAD stage and generates printer-ready G-code directly from images and part drawings. Instead of relying on an explicit 3D model, a hand-drawn or captured 2D image serves as the sole input. The framework first extracts slice-wise structural cues from the image and then employs a denoising diffusion probabilistic model (DDPM) over G-code sequences, parameterized by a one-dimensional convolutional network. Through iterative denoising, the model transforms Gaussian noise into coherent, executable print-move trajectories with corresponding extrusion parameters, establishing a direct and interpretable mapping from visual input to native toolpaths.
By producing structured G-code directly from 2D imagery, Image2Gcode eliminates the need for CAD or STL intermediates, lowering the entry barrier for additive manufacturing and accelerating the design-to-fabrication cycle. This approach supports low-overhead, on-demand prototyping from simple sketches or visual references and integrates naturally with upstream 2D-to-3D reconstruction modules to enable a fully automated pipeline from concept to physical artifact. The result is a flexible, computationally efficient framework that advances accessibility and responsiveness in design iteration, repair workflows, and distributed manufacturing contexts.

\end{abstract}

\section{Introduction}

Additive manufacturing (AM), commonly known as 3D printing, has democratized digital fabrication by enabling the direct conversion of virtual designs into physical co
mponents through layer-by-layer material deposition \cite{yadroitsev2021fundamentals}. Unlike traditional subtractive methods that shape parts by removing material from solid stock, AM fabricates structures additively, achieving intricate geometries often unattainable through conventional techniques \cite{faludi2015comparing, bacciaglia2024voxel, zhuang2023nurbs, elber2023vrepreview}, including topology-optimized components and lightweight lattice structures. This approach maximizes material efficiency while accelerating the design-to-production cycle, enabling rapid prototyping and the fabrication of complex, customized components \cite{jin2020machine}. These advantages have driven widespread AM adoption across diverse application domains over the past decade, including healthcare \cite{buj2021use, giannatsis2009additive, amaya-rivas2024future}, medical devices \cite{da2021comprehensive, haghiashtiani20203d}, aerospace \cite{najmon2019review}, and numerous other industries \cite{lacroix2023utilizing, stano2021additive, o2014advances}.

Among various AM techniques, Material Extrusion (MEX) has emerged as the most widely adopted process due to its low equipment cost, operational simplicity, and compatibility with diverse thermoplastic materials \cite{sapkota2024fdmflex, gibson2021additive}. MEX operates by extruding heated thermoplastic filament through a nozzle, depositing material layer-by-layer to construct parts from the bottom up. This accessibility has catalyzed the growth of open-source initiatives such as RepRap \cite{sells2010reprap, jones2011reprap} which have democratized 3D printing technology by enabling individuals, small businesses \cite{laplume2016open}, and research laboratories \cite{pearce2012building, pearce2013open} to develop and customize hardware, significantly expanding both the applications and reach of additive manufacturing \cite{staribratov2024steamedu}.

Despite these advances in hardware accessibility, the path from design concept to printed part remains constrained by a multi-stage workflow that demands specialized expertise at each step. The conventional AM pipeline begins with creating a 3D model in computer-aided design (CAD) software. This CAD model is then exported as a triangulated mesh file, which is subsequently processed by slicing software that divides the geometry into discrete horizontal layers. The slicer converts each layer into G-code, a low-level computer numerical control (CNC) programming language that specifies precise machine instructions for nozzle movement, material extrusion rates, and toolpath trajectories. These instructions are then transmitted directly to the 3D printer for physical fabrication \cite{pamidi2024practical,montalti2024cad}. Each stage in this pipeline introduces significant barriers: CAD modeling requires technical proficiency and familiarity with complex software interfaces, mesh processing demands understanding of resolution and file format considerations \cite{chen2024correlation}, slicing necessitates knowledge of print parameters and process optimization \cite{jadhav2024llm,vyavahare2020rpj,Kristiawan2021}, and iterative design refinement requires repeated cycles through the entire workflow \cite{macdonald20143d,pamidi2024practical}. This friction between creative vision and physical realization presents a substantial challenge, particularly for users lacking CAD expertise or those seeking rapid prototyping capabilities.

These constraints become particularly pronounced in application scenarios where the input is not a CAD model and the geometry is not explicitly specified. In such cases, the mandatory requirement for a valid 3D model undermines MEX's potential for quick, context-driven production, especially when target geometries are simple or highly situational \cite{Panico2025, Kartal2025}. In customized small-batch manufacturing, each new geometry necessitates complete workflow reconfiguration, creating throughput bottlenecks that severely limit scalability \cite{Daminabo2020}; educational and research laboratory environments face similar issues, where substantial time is spent on pre-processing activities (CAD modeling, file conversion, slicing configuration) than on the actual printing process itself \cite{nath2020optimizationAM}. These limitations collectively expose a critical paradox: the CAD-to-G-code workflow, despite its technical maturity and widespread adoption, has emerged as the primary barrier preventing MEX from achieving its potential as an agile.

Deep learning has emerged as a transformative approach, revolutionizing fields ranging from computer vision and natural language processing to scientific computing \cite{LeCun2015, Schmidhuber2015}. Unlike traditional methods that rely on predefined heuristics and rules, deep learning leverages large-scale datasets and hierarchical neural networks to automatically discover complex patterns and representations directly from data \cite{Silver2017, Brown2020}. In engineering design and manufacturing, deep learning has catalyzed a shift toward data-driven approaches, enabling systems to perform tasks that were once dependent on human expertise, such as design generation, fault detection, and optimization \cite{Silver2017, Brown2020}. Among the various deep learning approaches, generative models have gained particular attention for their ability to create novel designs and bypass traditional modeling workflows. Denoising Diffusion Probabilistic Models (DDPMs), in particular, have shown significant promise due to their stable training dynamics and capacity to generate high-quality, diverse samples. Unlike earlier models such as VAEs and GANs, DDPMs progressively transform noise into data through a reverse diffusion process, capturing fine-grained details and complex spatial relationships that are difficult for other generative models to resolve \cite{ho2020ddpm, song2020denoising}. Recently, DDPMs have been successfully applied to 3D geometry generation, showing remarkable results in synthesizing complex shapes and structures, with applications ranging from molecular modeling \cite{zhang2023diffmol} to 3D-printable mechanical and lattice structures \cite{jadhav2024generative}.



A particularly compelling direction is the use of deep learning to establish end-to-end learning frameworks for G-code generation. Rather than decomposing the workflow into isolated stages of slice, path planning, and parameter configuration, end-to-end models can be designed to bypass the CAD intermediate altogether, learning to map observable representations directly to executable machine instructions\cite{Liu2023, Panico2025}. This approach exploits the strengths of data-driven modeling, where patterns of effective nozzle trajectories and infill strategies are not explicitly prescribed, but instead emerge from the training distribution. The end-to-end paradigm reduces the cumulative errors introduced in intermediate steps, streamlines the workflow, and allows the system to optimize holistically for print quality, efficiency, and robustness\cite{Feng2022}.

\begin{figure}[t]
    \centering
    \includegraphics[width=1\linewidth]{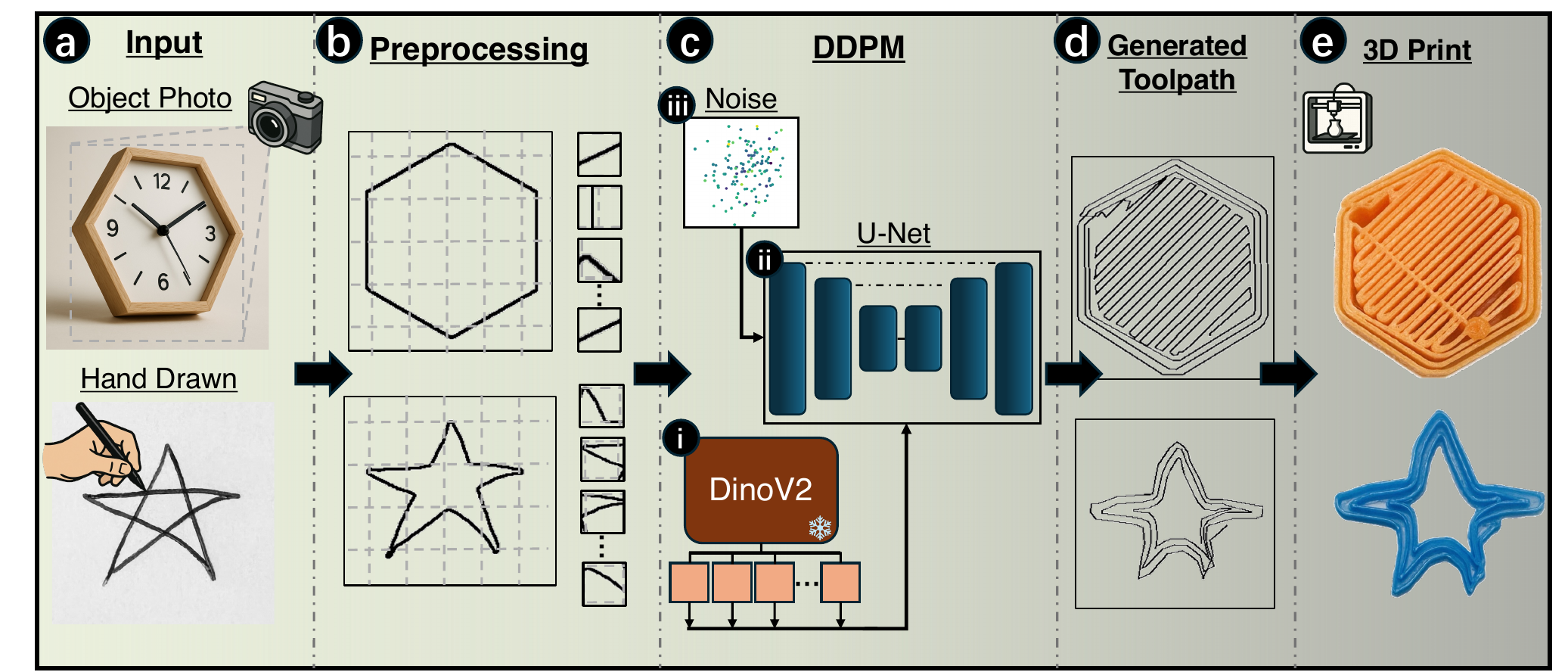}
    \caption{\textbf{Image2Gcode Overview.} Our end-to-end framework generates printer-ready G-code toolpaths directly from visual inputs. (a) The system accepts object photographs and hand-drawn sketches. (b) Preprocessing extracts geometric boundaries from input images. (c) A Denoising Diffusion Probabilistic Model (DDPM) comprises (i) a pre-trained DinoV2 vision encoder that extracts multi-scale semantic features, (ii) a 1D U-Net decoder conditioned on these features via cross-attention that progressively denoises sequences to generate toolpaths, and (iii) Gaussian noise initialization during inference. (d) The predicted G-code defines continuous extrusion trajectories capturing geometry-specific infill patterns. (e) Physical parts fabricated via MEX.}
    \label{fig.overview}
\end{figure}

Motivated by these challenges, we introduce Image2Gcode, an end-to-end framework that circumvents CAD representation by directly mapping visual inputs to executable manufacturing instructions. Our approach accepts diverse inputs including hand-drawn sketches and cross-sectional images, generating printer-ready material extrusion (MEX) toolpaths without intermediate 3D model reconstruction or conventional slicing operations. As shown in Figure~\ref{fig.overview}, our framework employs a Denoising Diffusion Probabilistic Model (DDPM) that integrates a pre-trained frozen DinoV2 \cite{oquab2023dinov2} vision encoder with a 1D U-Net decoder through multi-scale cross-attention. This design enables the U-Net to synthesize G-code sequences conditioned on hierarchical visual features extracted by DinoV2, progressively denoising from random initialization to coherent toolpaths. Critically, our data-driven formulation learns to predict the most probable infill patterns and extrusion strategies directly from observed manufacturing data rather than relying on hand-crafted heuristics. In its current form, Image2Gcode is applicable to 2D and 2.5D geometries, i.e., parts that can be represented by one or a small set of layer-wise consistent cross-sections. Although this imposes limitations for 3D parts with substantial layer to layer topological changes, the 2.5D regime covers a broad and highly practical subset of additive manufacturing. Moreover, establishing feasibility, physical constraints, and stability in the 2.5D setting constitutes an important step toward extending AI-driven, end-to-end generative manufacturing to truly complex 3D geometries. We demonstrate that this direct image-to-instruction paradigm achieves practical viability for geometrically constrained parts while significantly reducing computational overhead and workflow complexity compared to conventional CAD-dependent pipelines. The remainder of this paper details our methodology, experimental validation, and analysis of the framework's capabilities and limitations.

\section{Methodology}
The Image2Gcode framework comprises three primary components: a preprocessing module that extracts geometric features and trajectory sequences from sliced layers, a denoising diffusion probabilistic model that generates toolpath sequences conditioned on visual features, and a post-processing module that converts the generated sequences into validated, printer-ready G-code instructions.

\subsection{Dataset and Preprocessing}

\begin{figure}[H]
    \centering
    \includegraphics[width=1\linewidth]{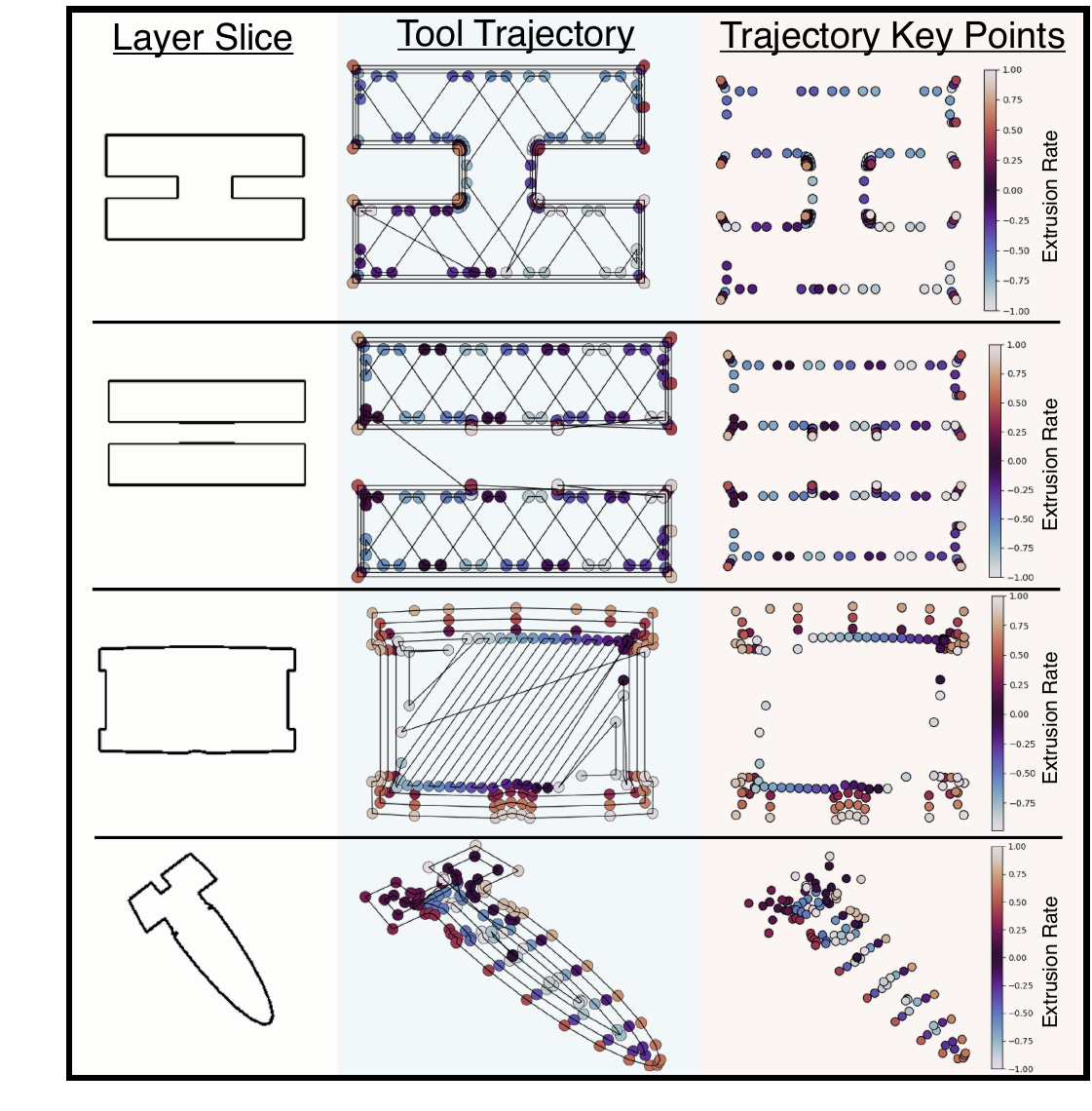}
    \caption{\textbf{Preprocessing pipeline.} Visualization illustrating the extraction of slice-level training pairs. For each layer (left column), the corresponding G-code toolpath is visualized with complete trajectories (middle column) and extracted key points colored by normalized extrusion rate (right column). Point colors encode the normalized extrusion values $E \in [-1, 1]$, where darker values indicate higher material deposition rates. This representation captures both spatial trajectory information through $(X, Y)$ coordinates and material deposition characteristics through the extrusion channel $E$.}
    \label{fig.2}
\end{figure}

The Image2Gcode framework is trained and evaluated using the Slice-100K dataset, a multimodal collection containing over 100,000 aligned STL-G-code pairs \cite{Jignasu2024slice100kdataset}. Each training sample comprises a single rendered slice image paired with its corresponding layer toolpath represented as trajectory keypoints, enabling supervised learning of the slice-to-toolpath mapping as illustrated in Figure~\ref{fig.2}.

The preprocessing pipeline extracts individual layers from each mesh-gcode pair to construct slice-level training data. For each STL file, the corresponding G-code is parsed to identify layer boundaries by detecting Z-axis changes in the toolpath. The parser detects coordinate system modes (e.g., \texttt{G90}/\texttt{G91} for positioning, \texttt{M82}/\texttt{M83} for extrusion) and converts all commands to a consistent absolute coordinate system. Motion commands \texttt{G1} are retained to extract $(X, Y, E)$ coordinates, where $X$ and $Y$ define the in-plane nozzle position and $E$ specifies the cumulative extrusion amount. When extruder reset commands \texttt{G92} are present, the cumulative extrusion offset is adjusted accordingly. Using the Z-coordinates extracted from the G-code, the corresponding STL mesh is sliced at identical heights using PyVista, ensuring precise alignment between each rendered slice image and its toolpath sequence. Each slice projected along the $+Z$ axis to produce grayscale images at $224 \times 224$ resolution.

For each layer, the extracted G-code forms an $N \times 3$ sequence $X_0 = \{(X_i, Y_i, E_i)\}_{i=1}^{N}$ representing the toolpath. Normalization is applied separately to the spatial and extrusion channels to account for their distinct geometric properties. The $(X, Y)$ coordinates are centered by the midpoint of the layer's bounding box and scaled by the larger dimension to preserve aspect ratio, mapping both channels to $[-1, 1]$. The extrusion values $E$ are first converted to absolute units and then normalized independently per layer using min-max scaling to $[-1, 1]$. This separate normalization strategy ensures that spatial geometry and material deposition are treated as distinct features during training. Sequences are padded or truncated to a fixed length $N_{\max}$ with zeros, and a binary mask $M \in \{0, 1\}^{N_{\max}}$ marks valid steps. 

This slice-based formulation decomposes 3D printing into independent 2D generation tasks, where the model learns to map cross-sectional geometry to toolpath keypoints on a per-layer basis. This strategy offers significant computational advantages, by operating on 2D slice representations rather than full volumetric data substantially reduces memory requirements and inference time, treating each layer as an independent sample dramatically increases data efficiency during training, and the per-layer formulation naturally accommodates variable layer heights and heterogeneous geometric complexity throughout the build without additional architectural modifications.

\subsection{Model Architecture}
\begin{figure}[H]
    \centering
    \includegraphics[width=1\linewidth]{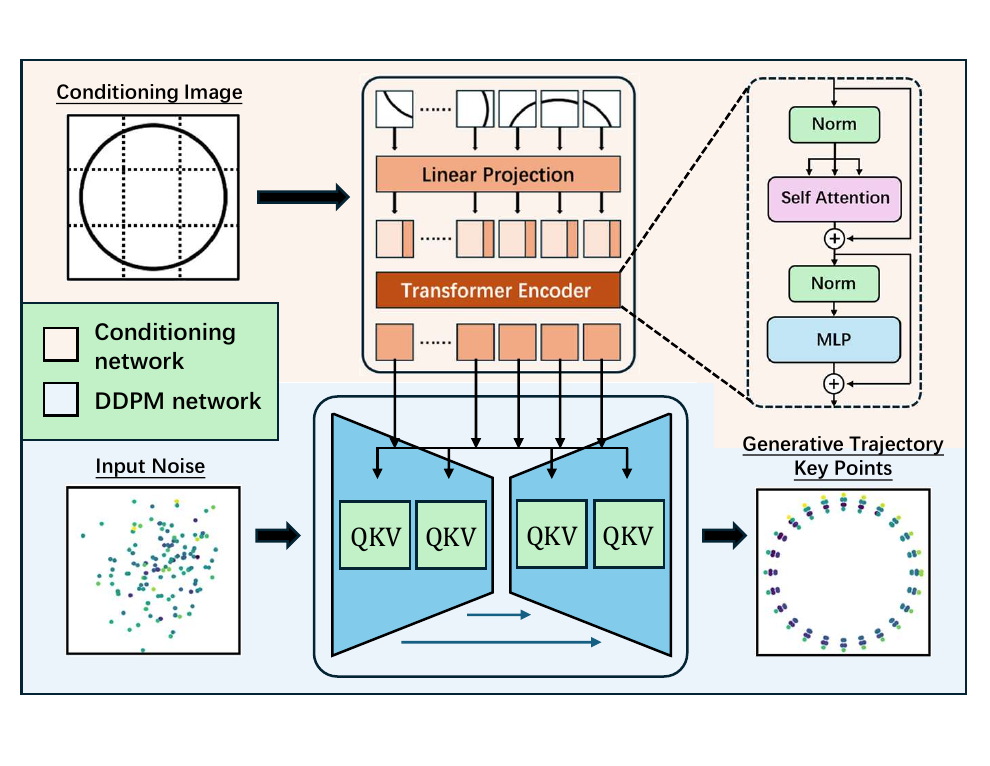}
    \caption{\textbf{Model Architecture.} The framework consists of a conditioning network and DDPM denoising network. The conditioning network processes input slice images through linear projection and a frozen DinoV2 transformer encoder to extract multi-scale visual features. The DDPM network uses a 1D U-Net with cross-attention mechanisms (QKV blocks) that fuse these visual features with trajectory sequences at multiple scales. The transformer block detail (right) shows the standard architecture with normalization, self-attention, and MLP layers connected via residual connections.}
    \label{fig.3}
\end{figure}
\subsubsection{Slice Encoder}

The visual encoder employs DinoV2~\cite{Oquab2023}, a self-supervised vision foundation model with a Vision Transformer (ViT) backbone~\cite{dosovitskiy2020image}, to extract hierarchical geometric features from slice images for conditioning the DDPM network. As illustrated in Figure~\ref{fig.3}, the input image is partitioned into fixed-size patches, with each patch linearly projected into a token embedding. Positional embeddings are added to preserve spatial information, and the resulting token sequence is processed by the transformer encoder to produce contextualized patch representations that capture both local geometric details and global structural relationships.
DinoV2 was trained on 142 million curated images using a student-teacher framework with multi-crop augmentation and center-temperature regularization, yielding robust visual representations without task-specific supervision. Leveraging these pretrained features provides several advantages for toolpath generation. First, DinoV2's representations inherently encode geometric primitives, edge structures, and spatial relationships that are directly relevant to interpreting slice geometry. Second, the pretrained features generalize effectively to out-of-distribution slice geometries not encountered during pretraining. Third, transfer learning dramatically reduces data requirements and training time compared to learning visual representations from scratch on the relatively smaller Slice-100K dataset.
The architecture uses the DinoV2-Small configuration with 14×14 patch size, 384-dimensional embeddings, and 12 transformer layers. For 224×224 input images, this produces a 16×16 spatial grid of 256 patch tokens that serve as multi-scale conditioning signals for the denoising network. The encoder is partially frozen during training, with only the final two transformer layers fine-tuned to adapt high-level features to the toolpath generation task while preserving the robust low-level geometric representations learned during pretraining.

\subsection{Denoising Diffusion Probabilistic Model}

The denoising diffusion probabilistic model serves as the generative core of the framework, learning to synthesize trajectory keypoints conditioned on the visual features extracted by the DinoV2 encoder. The model operates directly on the normalized keypoint representation $\tilde{X}_0 \in \mathbb{R}^{N_{\max} \times 3}$, where each keypoint consists of $(X, Y, E)$ coordinates. The input keypoint sequence is first embedded into a higher-dimensional feature space through a $1 \times 1$ convolution that projects each three-dimensional coordinate to $C_{\text{in}} = 128$ channels. Unlike many sequence models, no explicit per-step positional encoding is applied to the keypoints, as ordering information is inherently captured through the 1D convolutions, residual connections, and self-attention mechanisms within the U-Net backbone. The only sinusoidal embedding injected into the sequence pathway is the diffusion timestep $t$, which modulates all residual blocks to inform the network of the current noise level during the denoising process.

The forward diffusion process follows the standard DDPM formulation, gradually corrupting the clean keypoints $X_0$ with Gaussian noise over $T = 500$ timesteps according to $q(X_t | X_0) = \mathcal{N}(\sqrt{\bar{\alpha}_t} X_0, (1 - \bar{\alpha}_t) \mathbf{I})$, where $\bar{\alpha}_t = \prod_{s=1}^{t} \alpha_s$ controls the noise schedule\cite{ho2020ddpm}. Rather than predicting the noise term as in many diffusion implementations, the network directly reconstructs the clean keypoints $X_0$ from their noisy observation $X_t$ conditioned on slice features $c$. The training objective uses a masked L2 loss to account for variable-length keypoint sequences, defined as
\begin{equation}
\mathcal{L}_{\text{diff}} = \mathbb{E}_{X_0, t, c}\left[ \frac{1}{\sum_i M_i} \sum_{i=1}^{N_{\max}} M_i \left\| X_{0,i} - \hat{X}_{\theta,i}(X_t, t, c) \right\|_2^2 \right],
\end{equation}
where $c$ denotes the slice features from DinoV2, $\hat{X}_{\theta}(X_t, t, c)$ represents the predicted clean keypoints conditioned on both the noisy input and visual features, and the binary mask $M \in \{0,1\}^{N_{\max}}$ excludes padded positions from the loss computation\cite{karras2022edm}.

The generator architecture is a 1D U-Net with input channels of 128 and channel multipliers $(2, 2, 4, 6, 8)$ across five resolution scales. Each stage comprises residual blocks with GroupNorm and SiLU activations, strided convolutions for spatial downsampling in the encoder, and linear upsampling with skip connections in the decoder. The diffusion timestep $t$ is encoded via sinusoidal positional encoding, passed through a small MLP, and injected into each residual block through affine modulation (FiLM conditioning). This architecture was selected after systematic comparison of multiple channel configurations during early development.

The conditioning mechanism operates through multi-scale cross-attention layers integrated at multiple depths of the U-Net, enabling hierarchical fusion between visual features and keypoint representations. At each cross-attention layer, the keypoint features serve as queries $Q \in \mathbb{R}^{N \times d_q}$, while all $P = 256$ DinoV2 patch tokens provide keys and values $K, V \in \mathbb{R}^{P \times d_k}$. The attention mechanism computes
\begin{equation}
\text{Attention}(Q, K, V) = \text{softmax}\left(\frac{QK^\top}{\sqrt{d_h}}\right) V,
\end{equation}
with head dimension $d_h = 32$. The number of attention heads scales with the channel width at deeper layers to maintain computational efficiency \cite{vaswani2017attention}. This cross-attention design allows each keypoint in the sequence to attend to the entire spatial field of the slice image, injecting global geometric context while preserving the sequential structure through residual connections. Conditioning dropout is disabled to ensure deterministic alignment between visual features and generated keypoints during training.

\subsection{Post-Processing and G-code Generation}

During inference, the DDPM generates trajectory keypoints by starting from pure Gaussian noise and iteratively denoising over $T = 500$ steps using the same noise schedule as training. The model outputs a normalized keypoint sequence in the range $[-1, 1]$ with shape $[B, 3, L]$, where $B$ is the batch size, 3 represents the $(X, Y, E)$ channels, and $L$ is the sequence length. A predicted binary mask identifies valid keypoints, allowing padded positions to be excluded from subsequent processing.

The normalized representation provides significant flexibility for adapting generated toolpaths to different manufacturing contexts. Since the spatial coordinates $(X, Y)$ are normalized while preserving aspect ratio during preprocessing, the generated geometry can be uniformly scaled to arbitrary physical dimensions simply by adjusting the denormalization scale factor without requiring model retraining or affecting the structural integrity of the toolpath. Similarly, the normalized extrusion channel $E$ enables straightforward adaptation to different printer specifications. Extrusion parameters vary substantially across material extrusion systems due to differences in nozzle diameter, filament material properties, layer height settings, and extruder calibration. By storing the denormalization parameters separately, the same generated keypoint sequence can be scaled by printer-specific extrusion multipliers to accommodate these hardware variations, effectively decoupling the learned geometric relationships from machine-dependent material deposition rates.

The post-processing module performs denormalization using the cached parameters from preprocessing. The $(X, Y)$ coordinates are scaled by the layer bounding box dimensions and translated by the stored midpoint offset to restore physical positions in millimeters. The extrusion values $E$ are denormalized using the cached min-max parameters and then optionally scaled by an additional printer-specific multiplier to account for hardware characteristics. The denormalized keypoints are then formatted into standard G-code commands, with each keypoint $(X_i, Y_i, E_i)$ converted to a \texttt{G1} linear motion command of the form \texttt{G1 X\{X\_i\} Y\{Y\_i\} E\{E\_i\} F\{feedrate\}}. The feedrate parameter is set based on the target printer's recommended print speed. The sequence is prefixed with initialization commands including coordinate mode specification (\texttt{G90} for absolute positioning, \texttt{M83} for relative extrusion), temperature settings, and homing procedures, while standard shutdown commands including retraction and motor disabling are appended at the end. Basic validation checks ensure coordinates remain within physical build limits and verify monotonic extrusion progression throughout the layer, producing a complete single-layer toolpath ready for manufacturing execution.

\subsubsection{Implementation}
The model is implemented in PyTorch and trained across multiple GPUs with a per-GPU batch size of 64. Training runs for 800 epochs using the AdamW optimizer with initial learning rate $1 \times 10^{-4}$ and weight decay $10^{-2}$. The learning rate is adjusted via \texttt{ReduceLROnPlateau} scheduler monitoring validation loss with reduction factor 0.9, patience 20 epochs, and minimum rate $10^{-8}$. Gradient clipping is applied for stability, while mixed precision training and gradient checkpointing are disabled.

The diffusion model uses $x_0$-prediction with a cosine noise schedule over $T = 500$ timesteps. The reconstruction loss is mean squared error on $(X, Y, E)$ channels with fixed weights $[1.3, 1.3, 0.4]$ to account for their different physical scales, multiplied by SNR-based timestep weighting.

\section{Results and Discussion}

\begin{figure}[H]
    \centering
    \includegraphics[width=0.8\linewidth]{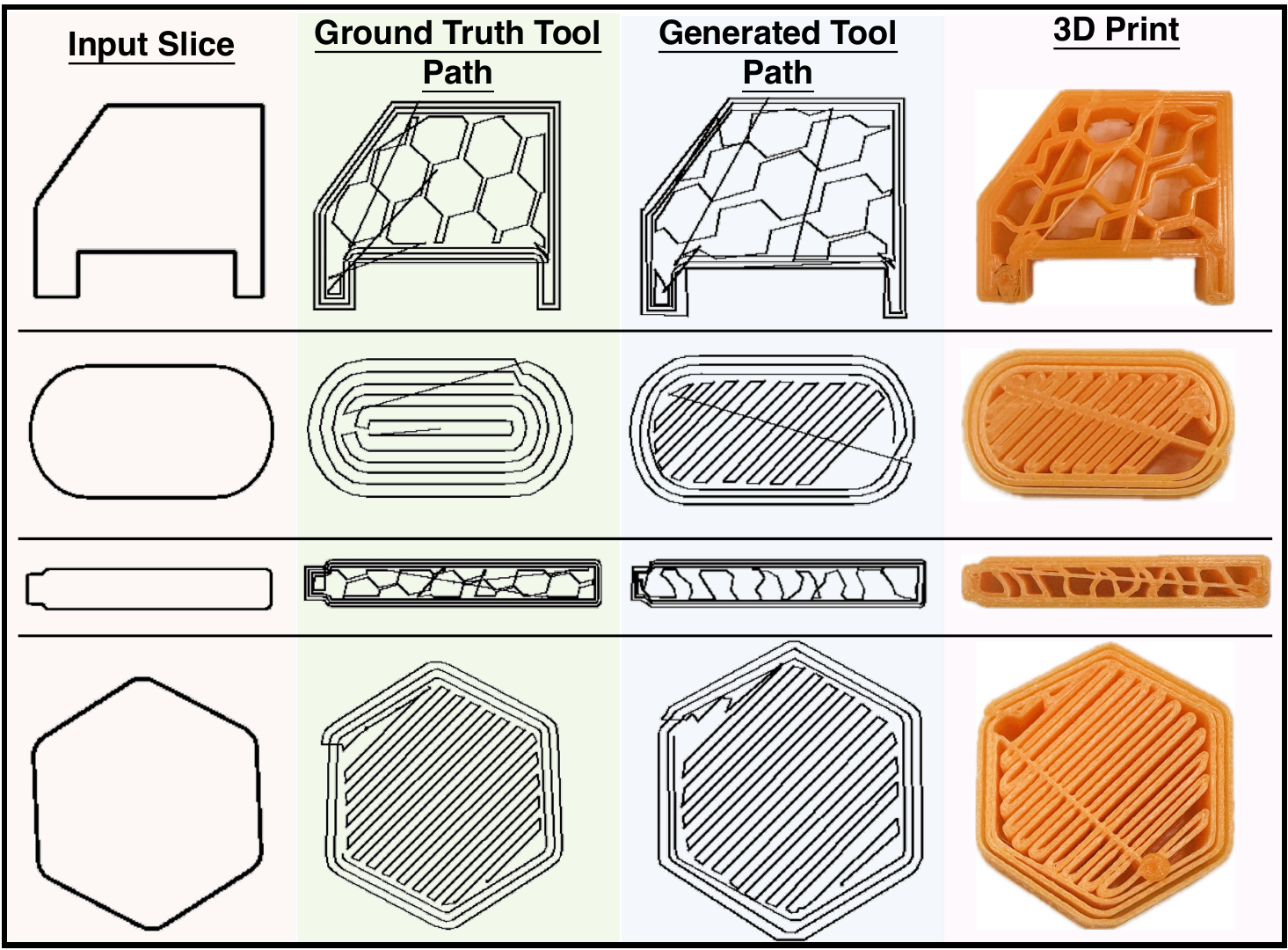}
    \caption{\textbf{Generated samples.} Qualitative results on validation samples from Slice-100K. For each geometry, the model generates structurally coherent toolpaths from input slice images. Generated toolpaths (third column) demonstrate accurate boundary reproduction and learned infill pattern selection, with physical prints (rightmost column) validating manufacturability and dimensional fidelity across diverse geometric primitives and infill strategies.}
    \label{fig:res_dataset}
\end{figure}
\begin{figure}[H]
    \centering
    \includegraphics[width=0.8\linewidth]{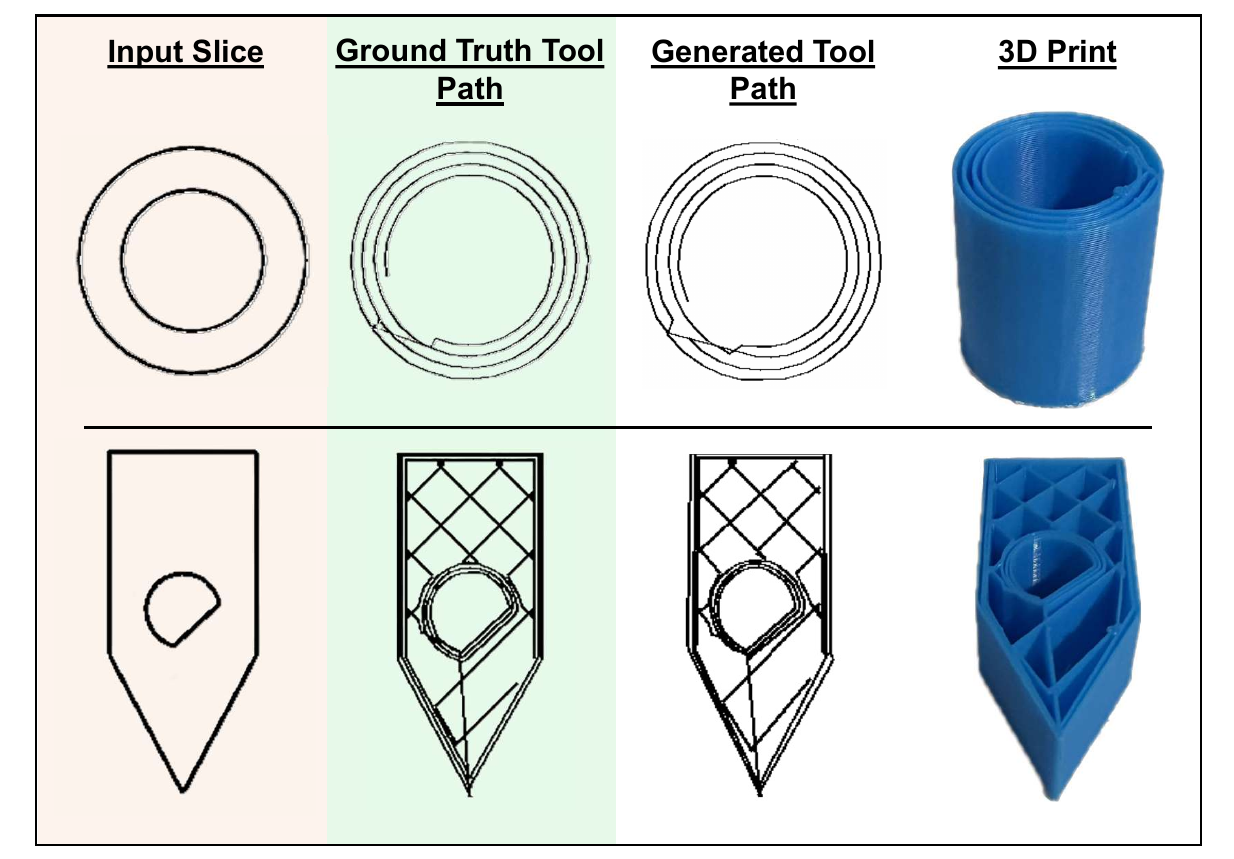}
    \caption{\textbf{Generated hollow samples.} The results from the Slice-100K validation sample. For the hollow geometries, the model can also generate structurally coherent tool paths from the input slice images. The generated tool paths (third column) and the physical prints (rightmost column) verify the manufacturability and dimensional fidelity when the geometric features of the input image are hollow.}
    \label{fig:res_hollow}
\end{figure}
\subsection{Generation Quality on Validation Data}

 To evaluate generation capabilities, inference is performed on the held-out 10\% validation split from Slice-100K. The model generates trajectory keypoints through iterative denoising over 500 timesteps, starting from pure Gaussian noise conditioned on input slice images. Figure~\ref{fig:res_dataset} presents representative results showing the input slice, ground-truth toolpath, generated toolpath, and corresponding physical print for each sample.

The generated toolpaths demonstrate strong geometric fidelity with accurately captured boundary geometry, well-closed perimeters, and faithful reproduction of sharp corners and geometric transitions. The model successfully infers diverse infill strategies from the training distribution, including honeycomb cellular structures, rectilinear line fill, organic patterns, and diagonal hatching that adapt to specific boundary geometry. These structures exhibit logical spatial organization with consistent orientation and appropriate density, reflecting learned trade-offs between material efficiency and structural integrity. Notably, the model also handles inputs containing hollow structures, preserving internal void regions while maintaining stable surrounding perimeters and context-appropriate internal patterning.

Physical fabrication validates the manufacturability of generated toolpaths. Printed parts maintain dimensional accuracy comparable to ground-truth designs with continuous surface quality and no layer delamination or severe under-extrusion. Minor discrepancies appear in high-curvature regions as slight variations in corner smoothing and local segment ordering, while occasional weaker bonding at perimeter-infill junctions suggests sensitivity to transient extrusion dynamics. These artifacts do not significantly compromise the overall structural integrity. Figure~\ref{fig:res_hollow} further highlights cases with hollow structures, where generated G-code produces clean void preservation and robust print outcomes consistent with the overall results.

The conditioning mechanism effectively bridges visual input and sequential generation, enabling the model to maintain global geometric constraints while selecting context-appropriate infill architectures without explicit rule-based programming. This demonstrates that the data-driven approach successfully learns the relationship between boundary geometry including hollow regions and manufacturable internal structures directly from observed examples.

\subsection{Generalization to Real-World Inputs}
\begin{figure}[H]
    \centering
    \includegraphics[width=1\linewidth]{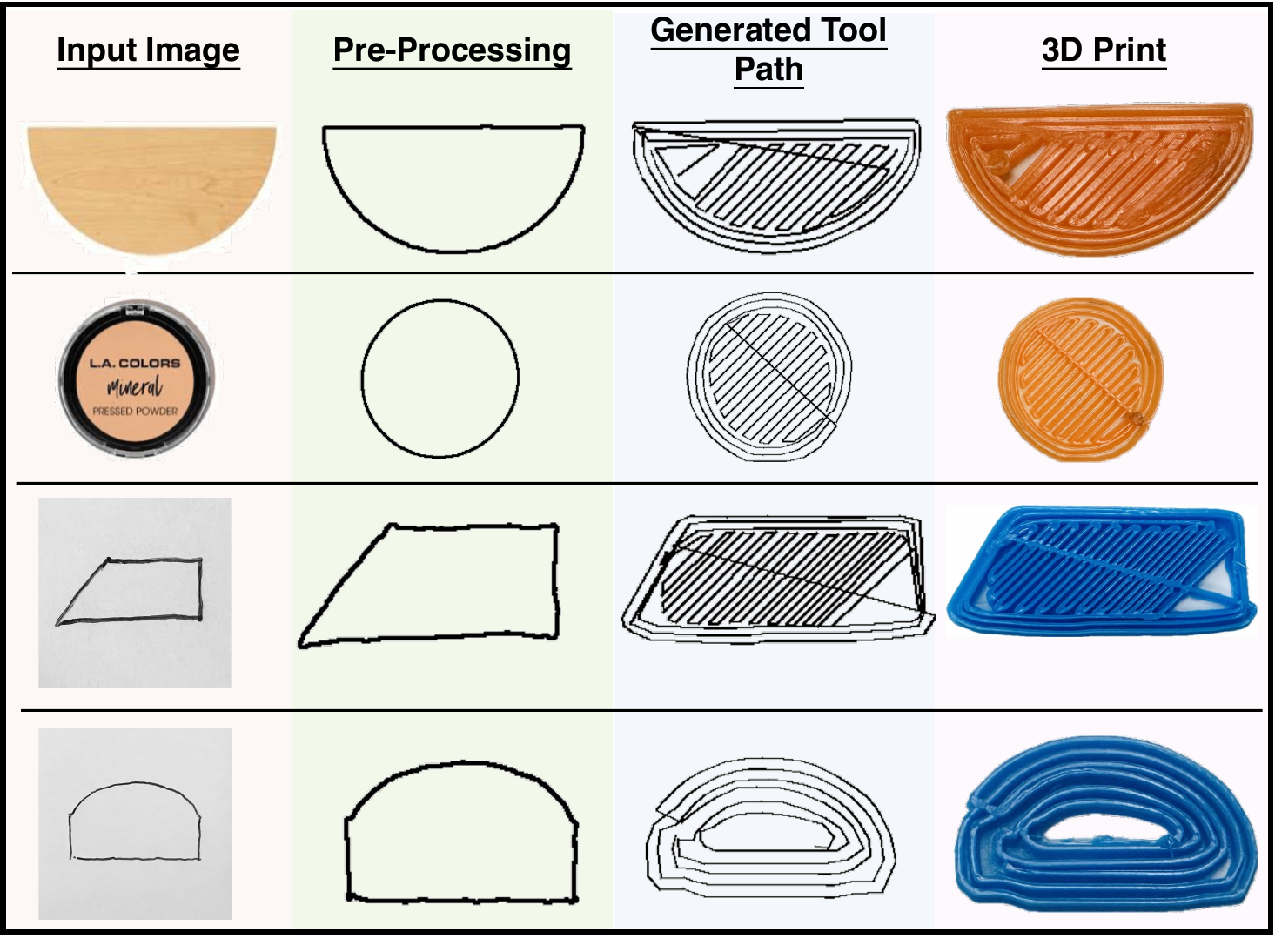}
    \caption{\textbf{Generalization to real-world inputs: } Photographs of physical objects (rows 1-2) and hand-drawn sketches (rows 3-4). The preprocessing module extracts boundary geometry from raw images, enabling the model to generate manufacturable toolpaths despite significant distribution shift from synthetic training data. Physical prints validate successful generalization across diverse input modalities.}
    \label{fig.6}
\end{figure}

To assess the model's ability to generalize beyond synthetic training data, inference is performed on photographs of physical objects and hand-drawn sketches. These inputs represent significant distribution shifts from the rendered slice images in Slice-100K, testing the robustness of the learned visual-to-toolpath mapping. Input images undergo preprocessing to extract boundary geometry through edge detection and contour extraction, producing binary silhouettes compatible with the model's expected input format. Figure~\ref{fig.6} presents results showing the original input, preprocessed boundary, generated toolpath, and corresponding physical print.

The model successfully generates manufacturable toolpaths from diverse real-world inputs including a photograph of a wooden bowl, a circular makeup compact, and hand-drawn sketches of geometric primitives. Despite the domain gap between photographic textures, lighting variations, hand-drawn irregularities, and the clean synthetic slices seen during training, the conditioning network extracts robust geometric features that enable coherent toolpath generation. The generated paths maintain well-defined perimeters that accurately follow the extracted boundaries while synthesizing contextually appropriate infill patterns including diagonal hatching and concentric shells.

Physical fabrication demonstrates practical viability across all test cases. The printed parts exhibit dimensional fidelity to the input geometry with structurally sound infill that provides mechanical integrity. The bowl and circular geometries print successfully despite their non-polygonal boundaries, while the hand-drawn sketches translate into clean geometric parts despite imperfect input linework. This generalization capability suggests that the pretrained DinoV2 encoder provides sufficiently abstract geometric representations to bridge the domain gap between synthetic training data and real-world visual inputs, enabling direct image-to-manufacturing workflows without retraining or domain adaptation.

\subsection{Novel Infill Pattern Generation}
\begin{figure}[H]
    \centering
    \includegraphics[width=1\linewidth]{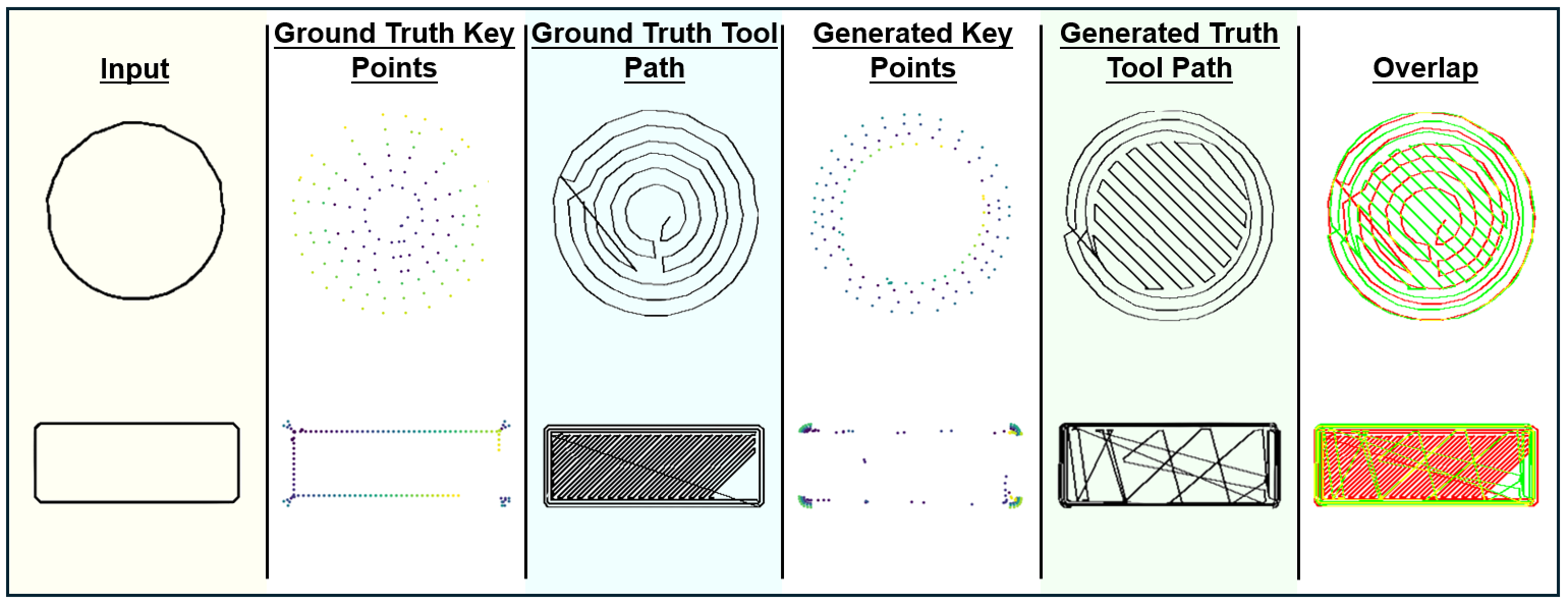}
    \caption{\textbf{Novel infill pattern.} Generation demonstrating diversity in the learned distribution. Ground truth toolpaths (columns 2-3) employ concentric and spiral strategies, while generated toolpaths (columns 4-5) produce alternative valid patterns including diagonal hatching and triangles architectures. The overlap visualization (column 6) shows ground truth in red and generated paths in green, revealing distinct topological approaches that achieve comparable spatial coverage.}
    \label{fig:novel_infill}
\end{figure}
Toolpath generation is inherently a many-to-one mapping problem where multiple valid infill strategies can satisfy the same geometric boundary constraints. To evaluate whether the model learns a distribution of valid solutions rather than memorizing specific training examples, we analyze cases where generated toolpaths deviate substantially from ground truth while maintaining structural validity. Figure~\ref{fig:novel_infill} presents comparative visualizations showing input geometry, ground truth keypoints and toolpath, generated keypoints and toolpath, and spatial overlap between the two strategies.

The results demonstrate that the model generates structurally coherent infill patterns that differ significantly from the ground truth training examples. For the circular geometry (row 1), the ground truth exhibits a concentric spiral pattern originating from the center, while the model generates a diagonal rectilinear hatching pattern with perimeter shells. For the rectangular geometry (row 2), the ground truth features a dense diagonal rectilinear hatching pattern, whereas the generated toolpath adopts a sparse triangular pattern. Despite these topological and density differences, both generated strategies produce manufacturable toolpaths with appropriate perimeter definition and effective area coverage.

The overlap visualization (rightmost column) reveals that while the spatial trajectories differ substantially in local path geometry, both strategies achieve comparable material deposition coverage within the boundary constraints. This behavior indicates that the diffusion model has learned the underlying geometric and manufacturing constraints rather than simply memorizing specific toolpath sequences from the training data. The ability to generate diverse valid solutions suggests the model captures a distribution over feasible infill strategies conditioned on boundary geometry, enabling adaptive pattern selection based on the learned trade-offs between printing efficiency, structural integrity, and geometric constraints present in the training distribution.

\subsection{Travel Distance Analysis}

\begin{figure}[H]
    \centering
    \includegraphics[width=1\linewidth]{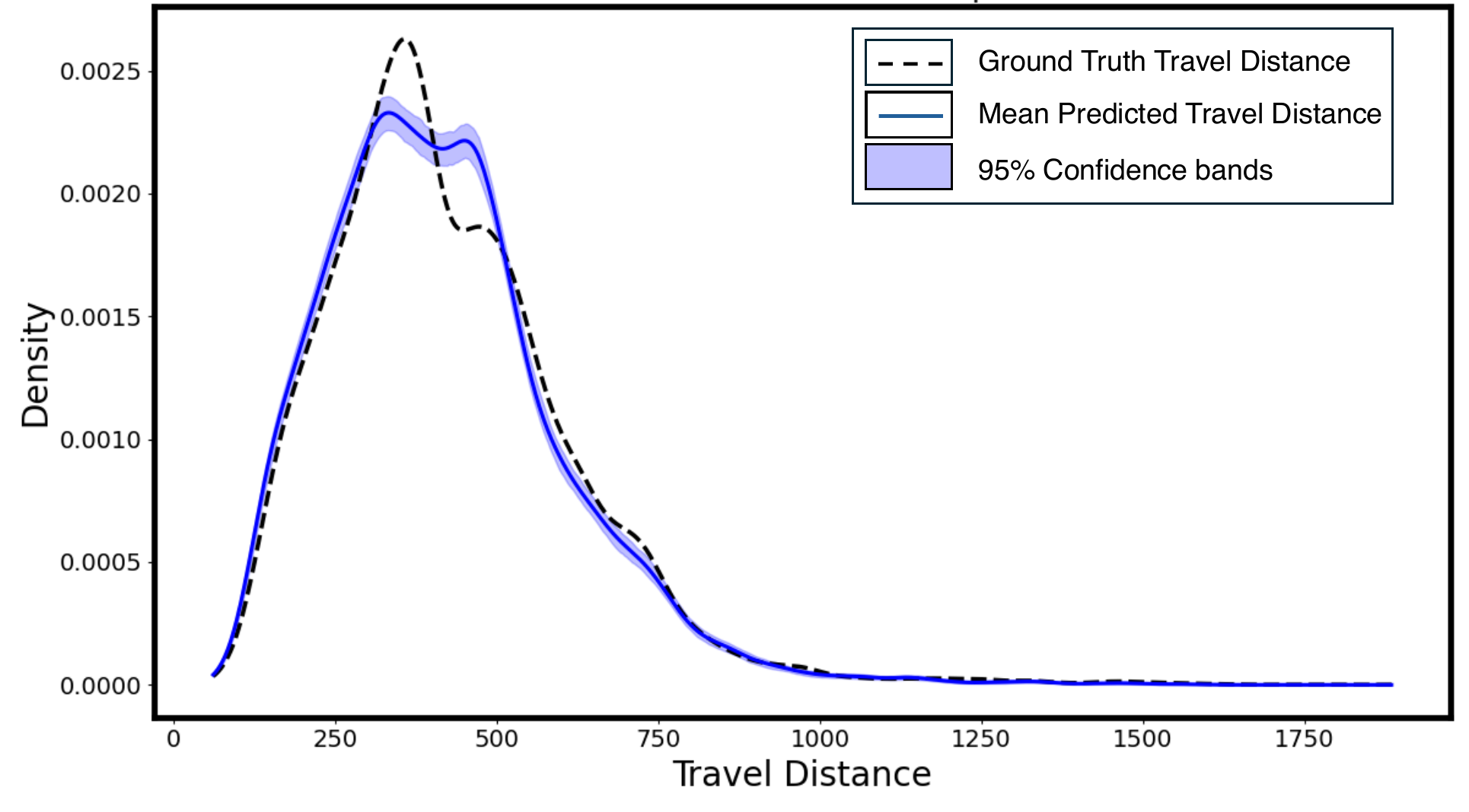}
    \caption{\textbf{Reduction in travel distance.} Kernel density estimation of travel distances comparing ground truth from heuristic slicers (dashed black) versus model predictions (solid blue) averaged across three independent runs. The predicted distribution shows a statistically significant 2.40\% reduction in mean travel distance while maintaining distributional alignment, indicating learned path efficiency. The narrow 95\% bootstrap confidence band demonstrates high generation consistency.}
    \label{fig:kde}
\end{figure}

To quantitatively assess toolpath efficiency, we analyze the distribution of total travel distances across validation samples. Travel distance represents the cumulative length of all nozzle movements during layer fabrication, encoding information about perimeter tracing, infill density, and path planning efficiency. Figure~\ref{fig:kde} presents kernel density estimates comparing ground truth travel distances from the Slice-100K dataset (generated by heuristic slicing algorithms) against mean predictions averaged across three independent generation runs, with 95\% bootstrap confidence intervals.

The distributions exhibit strong alignment with near-identical shapes and tail behavior, both peaking around 350-400 units. However, statistical analysis reveals a consistent leftward shift in the predicted distribution. The mean predicted travel distance is 417.73 units compared to 427.98 units for ground truth, representing a 2.40\% reduction. The narrow bootstrap confidence band indicates high reproducibility across independent runs, demonstrating that this reduction is consistent rather than stochastic.

This reduction suggests that the data-driven approach learns to generate more compact toolpath patterns while maintaining spatial coverage requirements. Since the model was trained on heuristic slicing outputs, the observed efficiency gain indicates the diffusion process captures underlying geometric regularities in the training distribution and synthesizes paths that achieve comparable coverage with reduced redundancy. The slight broadening of the predicted distribution relative to ground truth reflects the stochastic nature of diffusion sampling, which explores multiple valid solutions within the learned manifold.

Critically, successful physical fabrication of all generated toolpaths (as demonstrated in Figures~\ref{fig:res_dataset} and \ref{fig.6}) validates that this distance reduction does not compromise manufacturability, structural integrity, or geometric fidelity. The generated parts exhibit no evidence of insufficient coverage, weak bonding zones, or structural defects that might arise from overly aggressive path simplification. However, comprehensive mechanical testing would be required to fully characterize whether reduced travel distance translates to equivalent or improved structural performance compared to heuristic baselines. The observed efficiency represents a promising direction for data-driven toolpath optimization, suggesting that learned generative models can discover compact path planning strategies directly from manufacturing data.

\subsection{Limitations}

The current framework operates exclusively on 2D slice representations, limiting its application to 2D and 2.5D geometries, where the shape can be described by one or a small set of layer-wise consistent cross-sections. This slice-based formulation cannot directly handle complex 3D geometries with significant topological changes between layers, intricate internal cavities, or features requiring globally coordinated toolpath planning across multiple layers. Additionally, accuracy varies under elevated geometric complexity and fine-scale features, with measurable deviations arising from limited dataset diversity, computational constraints on model capacity, and the absence of explicit volumetric context that would inform interlayer relationships and global path consistency.

\section{Conclusion}

This work presents Image2Gcode, an end-to-end framework that generates printer-ready toolpaths directly from visual inputs without requiring intermediate CAD representations or rule-based slicing. By leveraging a denoising diffusion probabilistic model conditioned on pretrained visual features from DinoV2, the framework learns to synthesize manufacturable trajectory keypoints that capture both geometric boundary constraints and contextually appropriate infill strategies directly from observed manufacturing data. The slice-based formulation decomposes 3D printing into independent 2D generation tasks, reducing computational overhead while maintaining flexibility for variable layer heights and diverse geometric complexity. Experimental results demonstrate robust generalization from synthetic training data to real-world photographs and hand-drawn sketches, with physical fabrication validating the structural integrity and dimensional accuracy of generated toolpaths.

The proposed framework enables several practical use cases in rapid manufacturing workflows. For design iteration and prototyping, users can generate physical parts directly from concept sketches or reference photographs without CAD modeling expertise, significantly reducing the barrier to entry for additive manufacturing. In educational contexts, the direct image-to-part pipeline provides an intuitive interface for learning manufacturing principles without the overhead of traditional CAD-CAM toolchains. The normalized representation allows seamless adaptation to different printer configurations and material systems through simple scaling of spatial and extrusion parameters, facilitating deployment across heterogeneous manufacturing environments.

\section{Future Work}

Future work will focus on further expanding the Image2Gcode framework as a complement to traditional rule-based slicing, aiming to generate low-overhead toolpaths for 2D and 2.5D geometries derived directly from sketches or photographs, especially when CAD/STL models are unavailable. The goal is to integrate process rules and constraints into a learned generation pipeline: the specification of rules remains explicit (e.g., infill, surface priorities), while the realization of how these rules should adapt to geometry is learned from manufacturing data.

Future work will address current limitations through expanded data coverage, enhanced modeling capacity, and architectural extensions toward full 3D awareness. A hierarchical generation pipeline offers a promising direction wherein a coarse-level model first generates key cross-sections at critical Z-heights for a target 3D geometry, followed by the current slice-to-toolpath model operating on each layer independently. This decomposition would enable true 3D object fabrication while leveraging the computational efficiency of 2D generation at the layer level. Incorporating multiview or depth cues at the coarse stage, or adopting volumetric representations, could further improve cross-layer coherence and geometric fidelity.

Another key avenue is process-aware conditioning as a form of rule integration. In our framework, conditioning mechanisms should be viewed as the analog of slicer rule sets, providing explicit user- or application-specified manufacturing objectives and constraints, such as infill density and pattern type, material budget limits, or mechanical performance requirements. The key difference from conventional slicing lies in the implementation of the rules, and instead of relying solely on fixed heuristics, the model can learn geometry-adaptive manufacturing priors from data to better satisfy these constraints. Technically, such constraints can be encoded via additional input channels, structured tokens, or learned embeddings that modulate the diffusion process, enabling task-specific toolpath optimization while maintaining explicit controllability.

A particularly promising direction involves integration with autonomous manufacturing frameworks such as LLM-3D Print~\cite{jadhav2024llm}, which employs multi-agent LLM systems for defect detection, process monitoring, and adaptive parameter control in additive manufacturing. The current Image2Gcode framework generates toolpaths based solely on geometric conditioning, but incorporating natural language interfaces through LLMs would enable higher-level reasoning about manufacturing objectives. For instance, an LLM agent could interpret user specifications such as ``optimize for minimum print time while maintaining 80\% infill density'' or ``prioritize surface finish on top layer'' and translate these objectives into conditioning signals for the diffusion model. This could be implemented through learned embeddings that modulate the denoising process based on LLM-extracted manufacturing intent, or through iterative refinement where LLM agents evaluate generated toolpaths and provide feedback for regeneration. Such integration would bridge the gap between human design intent, automated visual-to-toolpath generation, and process-level optimization, creating a comprehensive AI-driven manufacturing pipeline from concept to fabrication.

Ultimately, these extensions toward full 3D awareness, explicit process parameter conditioning, and tighter integration with multi-agent manufacturing systems will further enhance the framework's versatility and robustness, advancing the vision of accessible, intelligent, and adaptive additive manufacturing workflows.

$$---------------------------------------$$
\bibliography{achemso-demo}

@book{yadroitsev2021fundamentals,
  title={Fundamentals of laser powder bed fusion of metals},
  author={Yadroitsev, Igor and Yadroitsava, Ina and Du Plessis, Anton and MacDonald, Eric},
  year={2021},
  publisher={Elsevier}
}

@article{faludi2015comparing,
  title={Comparing environmental impacts of additive manufacturing vs traditional machining via life-cycle assessment},
  author={Faludi, Jeremy and Bayley, Cindy and Bhogal, Suraj and Iribarne, Myles},
  journal={Rapid Prototyping Journal},
  volume={21},
  number={1},
  pages={14--33},
  year={2015},
  publisher={Emerald Group Publishing Limited}
}

@article{bacciaglia2024voxel,
  title   = {Voxel-based evolutionary topological optimization of connected structures for natural frequency optimization},
  author  = {Bacciaglia, A. and Ceruti, A. and Liverani, A.},
  journal = {International Journal of Mechanics and Materials in Design},
  volume  = {20},
  number  = {6},
  pages   = {1209--1228},
  year    = {2024},
  doi     = {10.1007/s10999-024-09722-8}
}

@article{zhuang2023nurbs,
  title        = {An efficient 2D/3D NURBS-based topology optimization implementation using page-wise matrix operation in MATLAB},
  author       = {Zhuang, C. and Xiong, Z. and Ding, H.},
  journal      = {Structural and Multidisciplinary Optimization},
  volume       = {66},
  articlenumber= {254},
  year         = {2023},
  doi          = {10.1007/s00158-023-03701-x}
}

@article{elber2023vrepreview,
  title        = {A Review of a B-spline based Volumetric Representation: Design, Analysis and Fabrication of Porous and/or Heterogeneous Geometries},
  author       = {Elber, Gershon},
  journal      = {Computer-Aided Design},
  volume       = {163},
  articlenumber= {103587},
  year         = {2023},
  month        = oct,
  doi          = {10.1016/j.cad.2023.103587}
}

@article{jin2020machine,
  title={Machine learning for advanced additive manufacturing},
  author={Jin, Zeqing and Zhang, Zhizhou and Demir, Kahraman and Gu, Grace X},
  journal={Matter},
  volume={3},
  number={5},
  pages={1541--1556},
  year={2020},
  publisher={Elsevier}
}

@article{buj2021use,
  title={Use of FDM technology in healthcare applications: recent advances},
  author={Buj-Corral, Irene and Tejo-Otero, Aitor and Fenollosa-Art{\'e}s, Felip},
  journal={Fused Deposition Modeling Based 3D Printing},
  pages={277--297},
  year={2021},
  publisher={Springer}
}

@article{giannatsis2009additive,
  title={Additive fabrication technologies applied to medicine and health care: a review},
  author={Giannatsis, J and Dedoussis, V},
  journal={The International Journal of Advanced Manufacturing Technology},
  volume={40},
  pages={116--127},
  year={2009},
  publisher={Springer}
}

@article{da2021comprehensive,
  title={A comprehensive review on additive manufacturing of medical devices},
  author={da Silva, Leonardo Rosa Ribeiro and Sales, Wisley Falco and Campos, Felipe dos Anjos Rodrigues and de Sousa, Jos{\'e} A{\'e}cio Gomes and Davis, Rahul and Singh, Abhishek and Coelho, Reginaldo Teixeira and Borgohain, Bhaskar},
  journal={Progress in Additive Manufacturing},
  volume={6},
  number={3},
  pages={517--553},
  year={2021},
  publisher={Springer}
}

@article{haghiashtiani20203d,
  title={3D printed patient-specific aortic root models with internal sensors for minimally invasive applications},
  author={Haghiashtiani, Ghazaleh and Qiu, Kaiyan and Zhingre Sanchez, Jorge D and Fuenning, Zachary J and Nair, Priya and Ahlberg, Sarah E and Iaizzo, Paul A and McAlpine, Michael C},
  journal={Science advances},
  volume={6},
  number={35},
  pages={eabb4641},
  year={2020},
  publisher={American Association for the Advancement of Science}
}

@article{najmon2019review,
  title={Review of additive manufacturing technologies and applications in the aerospace industry},
  author={Najmon, Joel C and Raeisi, Sajjad and Tovar, Andres},
  journal={Additive manufacturing for the aerospace industry},
  pages={7--31},
  year={2019},
  publisher={Elsevier}
}

@article{stano2021additive,
  title={Additive manufacturing aimed to soft robots fabrication: A review},
  author={Stano, Gianni and Percoco, Gianluca},
  journal={Extreme Mechanics Letters},
  volume={42},
  pages={101079},
  year={2021},
  publisher={Elsevier}
}

@article{o2014advances,
  title={Advances in three-dimensional rapid prototyping of microfluidic devices for biological applications},
  author={O'Neill, Paul F and Ben Azouz, Aymen and Vazquez, Mercedes and Liu, Jinghang and Marczak, Steven and Slouka, Zdenek and Chang, Hsueh Chia and Diamond, Dermot and Brabazon, Dermot},
  journal={Biomicrofluidics},
  volume={8},
  number={5},
  year={2014},
  publisher={AIP Publishing}
}

@article{montalti2024cad,
  title={From CAD to G-code: Strategies to minimizing errors in 3D printing process},
  author={Montalti, Andrea and Ferretti, Patrich and Santi, Gian Maria},
  journal={CIRP Journal of Manufacturing Science and Technology},
  volume={55},
  pages={62--70},
  year={2024},
  publisher={Elsevier}
}

@inproceedings{zhang2023diffmol,
  title={Diffmol: 3d structured molecule generation with discrete denoising diffusion probabilistic models},
  author={Zhang, Weitong and Wang, Xiaoyun and Smith, Justin and Eaton, Joe and Rees, Brad and Gu, Quanquan},
  booktitle={ICML 2023 Workshop on Structured Probabilistic Inference $\{$$\backslash$\&$\}$ Generative Modeling},
  year={2023}
}

@article{jadhav2024generative,
  title={Generative lattice units with 3d diffusion for inverse design: Glu3d},
  author={Jadhav, Yayati and Berthel, Joeseph and Hu, Chunshan and Panat, Rahul and Beuth, Jack and Barati Farimani, Amir},
  journal={Advanced Functional Materials},
  volume={34},
  number={41},
  pages={2404165},
  year={2024},
  publisher={Wiley Online Library}
}

@article{oquab2023dinov2,
  title={Dinov2: Learning robust visual features without supervision},
  author={Oquab, Maxime and Darcet, Timoth{\'e}e and Moutakanni, Th{\'e}o and Vo, Huy and Szafraniec, Marc and Khalidov, Vasil and Fernandez, Pierre and Haziza, Daniel and Massa, Francisco and El-Nouby, Alaaeldin and others},
  journal={arXiv preprint arXiv:2304.07193},
  year={2023}
}

@article{dosovitskiy2020image,
  title={An image is worth 16x16 words: Transformers for image recognition at scale},
  author={Dosovitskiy, Alexey},
  journal={arXiv preprint arXiv:2010.11929},
  year={2020}
}

@article{amaya-rivas2024future,
  title   = {Future trends of additive manufacturing in medical applications: An overview},
  author  = {Amaya-Rivas, Jorge L. and Perero, Bryan S. and Helguero, Carlos G. and Hurel, Jorge L. and Peralta, Juan M. and Flores, Francisca A. and Alvarado, Jos{\'e} D.},
  journal = {Heliyon},
  year    = {2024},
  month   = feb,
  volume  = {10},
  number  = {5},
  pages   = {e26641},
  doi     = {10.1016/j.heliyon.2024.e26641},
  url     = {https://doi.org/10.1016/j.heliyon.2024.e26641},
  publisher = {Elsevier}
}

@article{lacroix2023utilizing,
  title   = {Utilizing additive manufacturing and mass customization under capacity constraints},
  author  = {Lacroix, Rachel and Timonina-Farkas, Anna and Seifert, Ralf W.},
  journal = {Journal of Intelligent Manufacturing},
  year    = {2023},
  month   = jan,
  volume  = {34},
  number  = {1},
  pages   = {281--301},
  doi     = {10.1007/s10845-022-02007-x},
  url     = {https://doi.org/10.1007/s10845-022-02007-x},
  publisher = {Springer}
}

@incollection{sells2010reprap,
  title={RepRap: the replicating rapid prototyper: maximizing customizability by breeding the means of production},
  author={Sells, Ed and Bailard, Sebastien and Smith, Zach and Bowyer, Adrian and Olliver, Vik},
  booktitle={Handbook of Research in Mass Customization and Personalization: (In 2 Volumes)},
  pages={568--580},
  year={2010},
  publisher={World Scientific}
}

@article{jones2011reprap,
  title={RepRap--the replicating rapid prototyper},
  author={Jones, Rhys and Haufe, Patrick and Sells, Edward and Iravani, Pejman and Olliver, Vik and Palmer, Chris and Bowyer, Adrian},
  journal={Robotica},
  volume={29},
  number={1},
  pages={177--191},
  year={2011},
  publisher={Cambridge University Press}
}

@article{pearce2012building,
  title={Building research equipment with free, open-source hardware},
  author={Pearce, Joshua M},
  journal={Science},
  volume={337},
  number={6100},
  pages={1303--1304},
  year={2012},
  publisher={American Association for the Advancement of Science}
}

@book{pearce2013open,
  title={Open-source lab: how to build your own hardware and reduce research costs},
  author={Pearce, Joshua M},
  year={2013},
  publisher={Newnes}
}

@article{laplume2016open,
  title={Open-source, self-replicating 3-D printer factory for small-business manufacturing},
  author={Laplume, Andre and Anzalone, Gerald C and Pearce, Joshua M},
  journal={The International Journal of Advanced Manufacturing Technology},
  volume={85},
  pages={633--642},
  year={2016},
  publisher={Springer}
}

@book{gibson2021additive,
  title={Additive manufacturing technologies},
  author={Gibson, Ian and Rosen, David W and Stucker, Brent and Khorasani, Mahyar and Rosen, David and Stucker, Brent and Khorasani, Mahyar},
  volume={17},
  year={2021},
  publisher={Springer}
}

@article{sapkota2024fdmflex,
  title   = {A review on fused deposition modeling (FDM)-based additive manufacturing (AM) methods, materials and applications for flexible fabric structures},
  author  = {Sapkota, Ashok and Ghimire, Shree Kaji and Adanur, Sabit},
  journal = {Journal of Industrial Textiles},
  year    = {2024},
  volume  = {54},
  pages   = {1--51},
  doi     = {10.1177/15280837241282110},
  url     = {https://doi.org/10.1177/15280837241282110},
  publisher = {SAGE Publications}
}

@article{staribratov2024steamedu,
  title   = {3D technologies in STEAM education},
  author  = {Staribratov, Ivaylo and Manolova, Nikol},
  journal = {Discover Education},
  year    = {2024},
  month   = jul,
  volume  = {3},
  pages   = {92},
  doi     = {10.1007/s44217-024-00181-z},
  url     = {https://doi.org/10.1007/s44217-024-00181-z},
  publisher = {Springer Nature}
}

@article{nath2020optimizationAM,
  author  = {Nath, Paromita and Olson, Joseph D. and Mahadevan, Sankaran and Lee, Yung-Tsun T.},
  title   = {Optimization of fused filament fabrication process parameters under uncertainty to maximize part geometry accuracy},
  journal = {Additive Manufacturing},
  year    = {2020},
  volume  = {35},
  pages   = {101331},
  doi     = {10.1016/j.addma.2020.101331},
  publisher = {Elsevier}
}

@article{vyavahare2020rpj,
  author={Vyavahare, Siddharth and et al.},
  title={Fused deposition modelling: a review},
  journal={Rapid Prototyping Journal},
  year={2020},
  volume={26},
  number={1},
  pages={176--201},
  publisher={Emerald},
  doi={10.1108/RPJ-04-2019-0106}
}

@article{Kristiawan2021,
  author    = {Kristiawan, Ruben Bayu and Imaduddin, Fitrian and Ariawan, Dody and Ubaidillah and Arifin, Zainal},
  title     = {A review on the fused deposition modeling (FDM) 3D printing: Filament processing, materials, and printing parameters},
  journal   = {Open Engineering},
  year      = {2021},
  volume    = {11},
  number    = {1},
  pages     = {639--649},
  doi       = {10.1515/eng-2021-0063},
  publisher = {De Gruyter},
}

@article{Panico2025,
  author    = {Panico, Antonio and Corvi, Alberto and Collini, Luca and Sciancalepore, Corrado},
  title     = {Multi-objective optimization of FDM 3D printing parameters set via design of experiments and machine learning algorithms},
  journal   = {Scientific Reports},
  year      = {2025},
  volume    = {15},
  number    = {1},
  pages     = {16753},
  doi       = {10.1038/s41598-025-01016-z},
  publisher = {Nature Portfolio},
}

@article{Kartal2025,
  author    = {Kartal, Fatih and et al.},
  title     = {Mechanical performance optimization in FFF 3D printing using Taguchi design and machine learning approach with PLA/walnut shell composite},
  journal   = {Journal of Vinyl and Additive Technology},
  year      = {2025},
  volume    = {31},
  number    = {3},
  pages     = {622--638},
  doi       = {10.1002/vnl.22195},
  publisher = {John Wiley \& Sons Ltd},
}

@article{Daminabo2020,
  author    = {Daminabo, Samuel C. and Goel, Saurav and Grammatikos, Sotirios A. and Yazdani Nezhad, Hamed and Thakur, Vijay Kumar},
  title     = {Fused deposition modeling-based additive manufacturing (3D printing): techniques for polymer material systems},
  journal   = {Materials Today Chemistry},
  year      = {2020},
  volume    = {16},
  pages     = {100248},
  doi       = {10.1016/j.mtchem.2020.100248},
  publisher = {Elsevier},
}

@article{LeCun2015,
  title     = {Deep learning},
  author    = {LeCun, Yann and Bengio, Yoshua and Hinton, Geoffrey},
  journal   = {Nature},
  year      = {2015},
  volume    = {521},
  number    = {7553},
  pages     = {436--444},
  doi       = {10.1038/nature14539},
  publisher = {Nature Publishing Group},
}

@article{Schmidhuber2015,
  title     = {Deep learning in neural networks: An overview},
  author    = {Schmidhuber, J{\"u}rgen},
  journal   = {Neural Networks},
  year      = {2015},
  volume    = {61},
  pages     = {85--117},
  doi       = {10.1016/j.neunet.2014.09.003},
  publisher = {Elsevier},
}

@article{Silver2017,
  title     = {Mastering the game of Go without human knowledge},
  author    = {Silver, David and Schrittwieser, Julian and Simonyan, Karen and Antonoglou, Ioannis and Huang, Aja and Guez, Arthur and Hubert, Thomas and Baker, Lucas and Lai, Matthew and Bolton, Adrian and Chen, Yutian and Lillicrap, Timothy and Hui, Fan and Sifre, Laurent and van den Driessche, George and Graepel, Thore and Hassabis, Demis},
  journal   = {Nature},
  year      = {2017},
  volume    = {550},
  number    = {7676},
  pages     = {354--359},
  doi       = {10.1038/nature24270},
  publisher = {Nature Publishing Group},
}

@article{Brown2020,
  title     = {Language Models are Few-Shot Learners},
  author    = {Brown, Tom B. and Mann, Benjamin and Ryder, Nick and Subbiah, Melanie and Kaplan, Jared D. and Dhariwal, Prafulla and Neelakantan, Arvind and Shyam, Pranav and Sastry, Girish and Askell, Amanda and Agarwal, Sandhini and Herbert-Voss, Ariel and Krueger, Gretchen and Henighan, Tom and Child, Rewon and Ramesh, Aditya and Ziegler, Daniel M. and Wu, Jeffrey and Winter, Clemens and Hesse, Chris and Chen, Mark and Sigler, Eric and Litwin, Mateusz and Gray, Scott and Chess, Benjamin and Clark, Jack and Berner, Christopher and McCandlish, Sam and Radford, Alec and Sutskever, Ilya and Amodei, Dario},
  journal   = {Advances in Neural Information Processing Systems},
  year      = {2020},
  volume    = {33},
  pages     = {1877--1901},
  doi       = {10.48550/arXiv.2005.14165},
  publisher = {Curran Associates, Inc.},
}

@article{ho2020ddpm,
  author    = {Jonathan Ho and Ajay Jain and Pieter Abbeel},
  title     = {Denoising Diffusion Probabilistic Models},
  journal   = {arXiv preprint arXiv:2006.11239},
  year      = {2020},
  url       = {https://arxiv.org/abs/2006.11239},
  doi       = {10.48550/arXiv.2006.11239}
}

@article{song2020denoising,
  author    = {Jiaming Song and Chenlin Meng and Stefano Ermon},
  title     = {Denoising Diffusion Implicit Models},
  journal   = {arXiv preprint arXiv:2010.02502},
  year      = {2020},
  url       = {https://arxiv.org/abs/2010.02502},
  doi       = {10.48550/arXiv.2010.02502}
}

@article{Liu2023,
  title     = {Deep learning in additive manufacturing: From monitoring to design},
  author    = {Liu, Y. and Li, Y. and Cao, J.},
  journal   = {npj Computational Materials},
  year      = {2023},
  volume    = {9},
  number    = {1},
  pages     = {15},
  doi       = {10.1038/s41524-023-00995-5},
  publisher = {Nature Portfolio}
}

@article{Feng2022,
  title     = {Application of machine learning to optimize process parameters in fused deposition modeling of PEEK material},
  author    = {Feng, Qi and Maier, Walther and Möhring, Hans‐Christian},
  journal   = {Procedia CIRP},
  year      = {2022},
  volume    = {107},
  pages     = {1--8},
  doi       = {10.1016/j.procir.2022.04.001},
  publisher = {Elsevier}
}

@misc{Jignasu2024slice100kdataset,
  author    = {Anushrut Jignasu and Kelly O. Marshall and Ankush Kumar Mishra and Lucas Nerone Rillo and Baskar Ganapathysubramanian and Aditya Balu and Chinmay Hegde and Adarsh Krishnamurthy},
  title     = {Slice-100K: A Multimodal Dataset for Extrusion-based 3D Printing},
  year      = {2024},
  eprint    = {2407.04180},
  archivePrefix = {arXiv},
  note = {[dataset]}
}

@ARTICLE{Oquab2023,
  author = {Oquab, Maxime and Darcet, Timothee and Moutakanni, Theo and others},
  title = {{DINOv2}: Learning Robust Visual Features without Supervision},
  journal = {arXiv preprint arXiv:2304.07193},
  year = {2023},
  doi = {10.48550/arXiv.2304.07193},
}

@article{macdonald20143d,
  title={3D printing for the rapid prototyping of structural electronics},
  author={Macdonald, Eric and Salas, Rudy and Espalin, David and Perez, Mireya and Aguilera, Efrain and Muse, Dan and Wicker, Ryan B},
  journal={IEEE access},
  volume={2},
  pages={234--242},
  year={2014},
  publisher={IEEE}
}

@article{chen2024correlation,
  title={On the correlation between pre-processing workflow and dimensional accuracy of 3D printed parts in high-precision Material Jetting},
  author={Chen, Karin J and Elkaseer, Ahmed and Scholz, Steffen G and Hagenmeyer, Veit},
  journal={Additive Manufacturing},
  volume={91},
  pages={104335},
  year={2024},
  publisher={Elsevier}
}

@article{jadhav2024llm,
  title={Llm-3D print: large language models to monitor and control 3D printing},
  author={Jadhav, Yayati and Pak, Peter and Farimani, Amir Barati},
  journal={arXiv preprint arXiv:2408.14307},
  year={2024}
}

@article{pamidi2024practical,
  title={A practical guide to 3D printing for chemistry and biology laboratories},
  author={Pamidi, Arjun S and Spano, Michael B and Weiss, Gregory A},
  journal={Current Protocols},
  volume={4},
  number={10},
  pages={e70036},
  year={2024},
  publisher={Wiley Online Library}
}

@article{c,
  title={Real-time tool-path planning using deep learning for subtractive manufacturing},
  author={Feng, Yi-fei and Ma, Hong-Yu and Shen, Li-Yong and Yuan, Chun-Ming and Jiang, Xin},
  journal={IEEE Transactions on Industrial Informatics},
  volume={20},
  number={4},
  pages={5979--5988},
  year={2023},
  publisher={IEEE}
}

@inproceedings{karras2022edm,
  title     = {Elucidating the Design Space of Diffusion-Based Generative Models},
  author    = {Karras, Tero and Aittala, Miika and Aila, Timo and Laine, Samuli},
  booktitle = {Advances in Neural Information Processing Systems (NeurIPS)},
  year      = {2022},
  eprint    = {2206.00364},
  archivePrefix = {arXiv},
}

@inproceedings{vaswani2017attention,
  title     = {Attention Is All You Need},
  author    = {Vaswani, Ashish and Shazeer, Noam and Parmar, Niki and Uszkoreit, Jakob and Jones, Llion and Gomez, Aidan N. and Kaiser, {\L}ukasz and Polosukhin, Illia},
  booktitle = {Advances in Neural Information Processing Systems (NeurIPS)},
  year      = {2017}
}

\end{document}